\documentclass[preprint,12pt]{elsarticle}

\usepackage{amssymb}
\usepackage{amsmath}
\usepackage{booktabs}

\usepackage{graphicx}  
\usepackage{multirow}  
\usepackage{bbm}       
\usepackage{xcolor}    

\begin{document}

\begin{frontmatter}




\title{An Interpretable Local Editing Model for Counterfactual Medical Image Generation}

\author{Hyungi Min, Taeseung You, Hangyeul Lee, Yeongjae Cho, Sungzoon Cho\thanks{Corresponding author: zoon@snu.ac.kr}}

\affiliation{organization={Seoul National University},
            addressline={1 Gwanak-ro, Gwanak-gu},
            city={Seoul},
            postcode={08826},
            state={},
            country={Republic of Korea}}

\begin{abstract}
Counterfactual medical image generation have emerged as a critical tool for enhancing AI-driven systems in medical domain by answering "what-if" questions. However, existing approaches face two fundamental limitations: First, they fail to prevent unintended modifications, resulting collateral changes in demographic attributes when only disease features should be affected. Second, they lack interpretability in their editing process, which significantly limits their utility in real-world medical applications. To address these limitations, we present \textit{InstructX2X}, a novel interpretable local editing model for counterfactual medical image generation featuring \textit{Region-Specific Editing}. This approach restricts modifications to specific regions, effectively preventing unintended changes while simultaneously providing a \textit{Guidance Map} that offers inherently interpretable visual explanations of the editing process. Additionally, we introduce \textit{MIMIC-EDIT-INSTRUCTION}, a dataset for counterfactual medical image generation derived from expert-verified medical VQA pairs. Through extensive experiments, InstructX2X achieve state-of-the-art performance across all major evaluation metrics. Our model successfully generates high-quality counterfactual chest X-ray images along with interpretable explanations.
\end{abstract}

\begin{graphicalabstract}
\end{graphicalabstract}

\begin{highlights}
\item We propose a novel interpretable local editing model, InstructX2X, which effectively addresses existing challenges in counterfactual medical image generation.
\item We introduce innovative region-specific editing technique to ensure precisely controlled modification and enhance interpretability by providing guidance map.
\item We release \textit{MIMIC-EDIT-INSTRUCTION}, a high-quality instruction-based editing dataset for counterfactual medical image generation, constructed from expert-verified medical VQA pairs and enriched with location and severity information to support precise, localized edits and reproducible preprocessing.
\item InstructX2X demonstrates state-of-the-art performance through extensive experiments.
\end{highlights}


\end{frontmatter}




\section{Introduction}
\begin{figure} 
    \centering 
    \includegraphics[width=\textwidth]{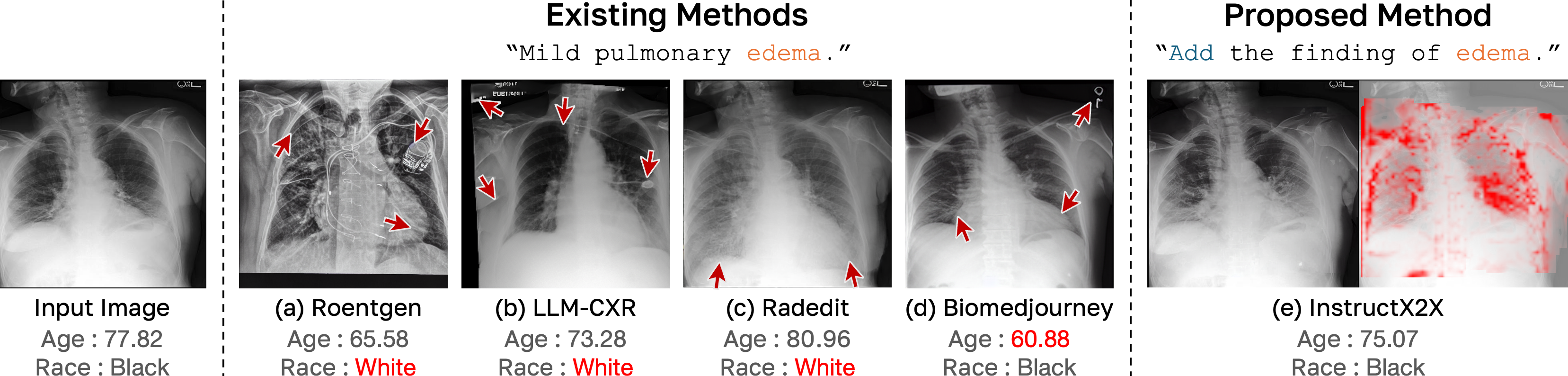} 
    \caption{Comparison of counterfactual medical image generation results between existing methods and our proposed approach. When adding edema features to an input chest X-ray image, existing methods (a–d) demonstrate unintended modifications (red arrows), causing significant variations in age and race (note the demographic predictions below each image). In contrast, InstructX2X preserves the demographic attributes while achieving precise editing and provides a visual explanation via guidance map (red overlay).}
    \label{comparison}
\end{figure}

Counterfactual medical image generation is an emerging approach that enhances AI-driven high-stakes medical decision-making. This methodology aims to answer what-if questions such as "How would this medical image change if the patient had a different disease?"~\cite{gifsplanation,attrinet}. By precisely manipulating target features in medical image while preserving unrelated attributes, this technique generates realistic edited images and helps uncover causal structures or biases in AI models. Counterfactual medical image generation offers various applications, such as evaluating model robustness~\cite{radedit}, providing counterfactual explanations~\cite{gifsplanation,dreamr,singla2023explaining}, enhancing classifier performance~\cite{attrinet,ktena2024generative}, and detecting anomalies~\cite{fontanella2024diffusion,sanchez2022healthy}.

Despite promising applications of counterfactual medical image generation, several technical challenges remain unresolved. A critical issue is the unintended modification of unrelated attributes when manipulating target features. In the context of chest radiography, Figure~\ref{comparison} illustrates such failure cases: when adding the edema feature to the input chest X-ray image, methods (a-c) alter the racial characteristics, while method (d) significantly changes the age attribute, despite the fact that these demographic attributes are independent of the edema features. Such unintended modifications distort the original clinical presentation and compromise the validity of the generated images ~\cite{xia2024mitigating}.




Another critical challenge in counterfactual medical image generation is a lack of interpretability. Interpretability (such as visual explanation) helps users to understand the model's decision-making process and validate the appropriateness of modifications~\cite{gifsplanation,amann2020explainability}. Current methods predominantly adopt post-hoc explanation techniques for model interpretation~\cite{roentgen,biomedjourney}. Although visually compelling, recent studies have demonstrated that these explanations frequently fail to represent the true decision mechanisms of the underlying models~\cite{han2022evaluating,white2020measurable,rudin2019stop,attrinet}. Such unreliable interpretability severely restricts the utility of counterfactual images for both clinical applications and model evaluation~\cite{borys2023explainable,rudin2019stop,amann2020explainability}.

To address these two critical limitations, we propose \textbf{InstructX2X}, a novel interpretable local editing model for counterfactual medical image generation. Our model introduces a \textit{Region-Specific Editing} approach that restricts editing to specific regions, preventing unintended modifications. Our targeted editing mechanism excludes potential spurious correlations outside the region of interest, resulting in highly reliable counterfactual images. Additionally, our region-specific editing methodology provides a \textit{Guidance Map}, visualized as the red overlay in Figure~\ref{comparison}(e), which highlights the modified areas, offering clear visual explanations of how the model processes the editing instructions. InstructX2X achieves inherent interpretability by directly revealing the decision mechanism to users, eliminating the need for post-hoc explanations of uncertain reliability.

Furthermore, the development of reliable counterfactual medical image generation has been constrained by the scarcity of datasets with reliable editing descriptions. To overcome this data deficiency, we repurpose an existing dataset from a different task domain into \textit{MIMIC-EDIT-INSTRUCTION}, a new counterfactual medical image generation dataset. We leverage expert-verified medical VQA pairs, unlike existing approaches that depend on large language models to generate editing descriptions without clinical validations~\cite{biomedjourney,Instructpix2pix}.





The key contributions of our research are:
\begin{enumerate}
    \item We propose a novel interpretable local editing model, InstructX2X, which effectively addresses existing challenges in counterfactual medical image generation.

    \item We introduce innovative region-specific editing technique to ensure precisely controlled modification and enhance interpretability by providing guidance map.

    \item We release \textit{MIMIC-EDIT-INSTRUCTION}, instruction-based editing dataset for counterfactual medical image generation derived from expert-verified medical VQA pairs.
    
    \item InstructX2X demonstrates state-of-the-art performance through extensive experiments.
    
    \end{enumerate}

\section{Method}
\begin{figure}[t]
    \centering
    \includegraphics[width=\textwidth]{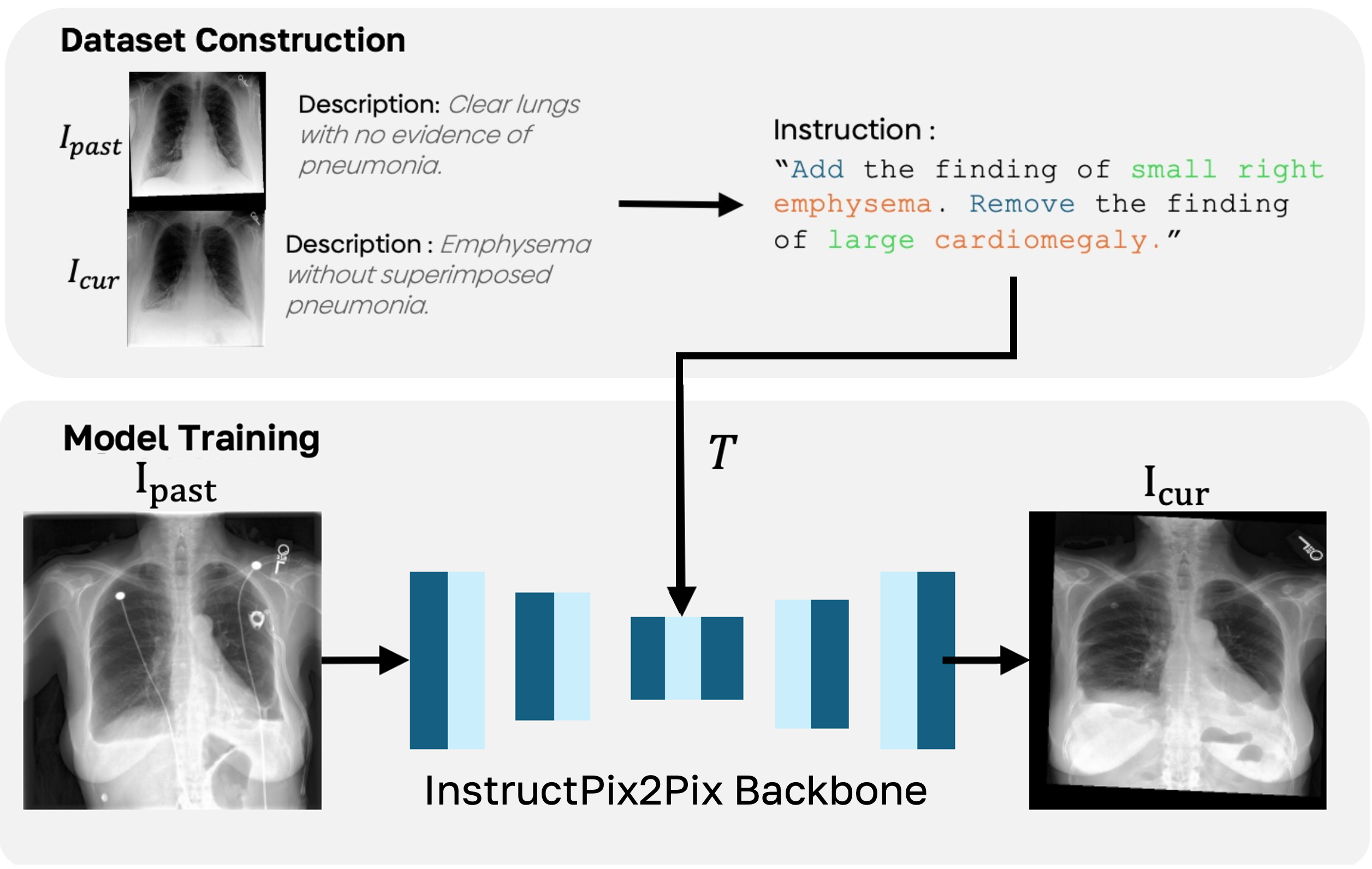}
    \caption{Overview of the InstructX2X training framework. The pipeline is adapted from the InstructPix2Pix architecture~\cite{Instructpix2pix}, modified to accept longitudinal chest X-ray pairs. The \textbf{top panel} illustrates the dataset construction process converting descriptions pairs into the \textit{MIMIC-EDIT-INSTRUCTION} data. The \textbf{bottom panel} shows the training pipeline where the model learns to transform $I_{past}$ to $I_{cur}$ using the constructed instructions.}
    \label{fig:training_framework}
\end{figure}

In this chapter, we introduce \textit{InstructX2X}, our instruction-based framework for counterfactual medical image editing. We describe the construction of the \textit{MIMIC-EDIT-INSTRUCTION} dataset and present our \textit{Region-Specific Editing} mechanism together with its visual representation, the \textit{Guidance Map}. The overall training pipeline of InstructX2X is illustrated in Figure~\ref{fig:training_framework}, and the inference procedure based on Region-Specific Editing is shown in Figure~\ref{fig:rse_mechanism}.

\subsection{MIMIC-EDIT-INSTRUCTION}

\subsubsection{Dataset preparation}
\label{Dataset}
InstructX2X utilizes three publicly available datasets: MIMIC-CXR~\cite{mimic-cxr}, MIMIC-Diff-VQA~\cite{mimicdiffvqa}, and MS-CXR~\cite{MS-CXR}. MIMIC-CXR contains 377,110 chest X-ray images and 227,827 radiology reports from 63,478 patients, while MIMIC-Diff-VQA builds upon it with 164,324 pairs of longitudinal chest X-rays and 700,703 expert-verified question-answer pairs. MS-CXR provides phrase grounding with 1,162 radiologist-annotated image-sentence pairs across eight diseases. From MIMIC-Diff-VQA, pair of images are selected and resized to $512 \times 512$.

\subsubsection{}ection{View Filtering and Image Registration}
To ensure the consistency of anatomical structures suitable for local editing, we specifically target Posterior-Anterior (PA) view images. We filter the dataset to retain only PA views using metadata and an external X-ray view classifier~\cite{xinario2015chestviewsplit}.

Since the input and target images in longitudinal pairs may have different patient positionings, precise alignment is essential. We perform rigid registration using SimpleITK~\cite{lowekamp2013}. To improve registration robustness against noise, we apply Bilateral Filtering with a domain sigma of 2.0 and a range sigma of 50.0 before alignment. We restrict geometric transformations to rotation, scaling, and translation to preserve pathological features. Finally, to guarantee high-quality alignment, we evaluate the registration using Mutual Information (MI) scores. We discard pairs with an MI score higher than -0.88, ensuring that only well-aligned pairs are used for training.

\subsubsection{Instruction Generation}
\label{Instruction Construction}

Existing approaches often use LLMs to generate editing descriptions~\cite{biomedjourney,Instructpix2pix}, but the lack of expert validation may result in clinical inaccuracies. To address this issue, we repurpose the MIMIC-Diff-VQA dataset, which provides expert-verified descriptions of temporal changes in chest X-rays with a 97.33\% validation rate. This dataset serves as a reliable source for constructing image-editing instructions—beyond its original VQA purpose. We identified three core operations that form an intuitive instruction of medical image modifications:

\begin{itemize}
\item \textbf{Add}: Introducing new findings or symptoms.
\item \textbf{Remove}: Eliminating existing findings or symptoms.
\item \textbf{Change the level}: Adjusting the severity level of present abnormalities.
\end{itemize}
As illustrated in top panel of Figure~\ref{fig:training_framework}, this expert-verified approach eliminates the need for LLMs in instruction construction. Beyond simple class labels, we extract detailed metadata including \textbf{Anatomical Location} (e.g., left lower lobe, right base) and \textbf{Severity Information} (e.g., mild, moderate, severe) using a rule-based extraction algorithm from the text. This structured approach allows for more precise and controllable editing. By decomposing complex medical changes into instructions, we establish \textit{MIMIC-EDIT-INSTRUCTION}, a new dataset for counterfactual medical image generation that maintains clinical precision while providing more precise and intuitive control~\cite{Instructpix2pix}.
 



\subsubsection{Dataset Statistics}
The final \textit{MIMIC-EDIT-INSTRUCTION} dataset consists of 21{,}957 high-quality samples.
We follow the official MIMIC-CXR patient-wise split and use patients in the P19 folder as a holdout cohort.
Specifically, we assign 19{,}204 samples to the training set and 2{,}248 samples to the holdout set
(P19 patients), and further divide the remaining data into a test set (334 samples) and a validation
set (171 samples) for model selection and hyperparameter tuning.
The detailed statistics of the dataset splits and the distribution of editing operations are
summarized in Table~\ref{table:dataset_stats}.

Regarding the editing operations, the dataset contains a total of 29{,}197 operations across
21{,}957 samples, corresponding to an average of approximately 1.33 operations per sample.
The distribution is relatively balanced between \textit{Add} (14{,}195; 48.6\%) and
\textit{Remove} (14{,}172; 48.5\%), while \textit{Change the level} accounts for a smaller
portion (830; 2.8\%).
The larger number of operations compared to samples reflects that a single longitudinal
pair may involve multiple simultaneous changes (e.g., adding one finding while removing another),
which is important for modeling clinically realistic disease trajectories.

\begin{table}[h]
\centering
\caption{Statistics of the MIMIC-EDIT-INSTRUCTION dataset. The dataset is split into training, holdout, test, and validation sets. The operation distribution shows the count of each editing type (Add, Remove, Change). Note that the total count of operations exceeds the number of samples because some samples contain multiple operations.}
\label{table:dataset_stats}
\renewcommand{\arraystretch}{1.2}
\setlength{\tabcolsep}{10pt}
\begin{tabular}{l|r}
\toprule
\textbf{Split} & \textbf{Number of Samples} \\
\hline
Train & 19,204 \\
Holdout & 2,248 \\
Test & 334 \\
Validation & 171 \\
\hline
\textbf{Total} & \textbf{21,957} \\
\bottomrule
\toprule
\textbf{Operation Type} & \textbf{Count (Percentage)} \\
\hline
Add & 14,195 (48.5\%) \\
Remove & 14,172 (48.4\%) \\
Change & 830 (2.8\%) \\
\hline
\textbf{Total Operations} & \textbf{29,197} \\
\bottomrule
\end{tabular}
\end{table}

\subsection{Region-Specific Editing}
\begin{figure}[!t]
    \centering
    \includegraphics[width=\textwidth]{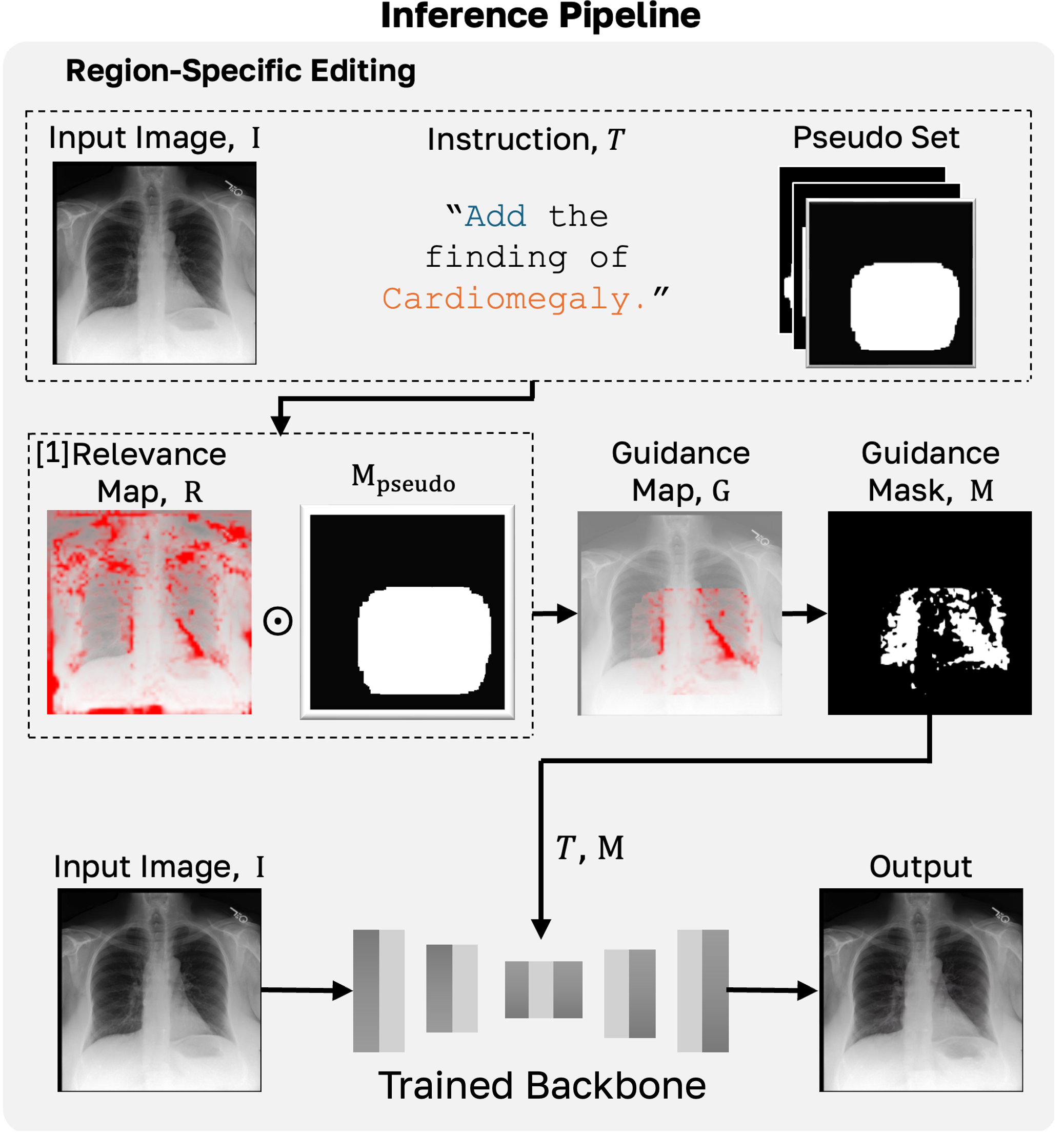}
    \caption{Overview of our proposed Region-Specific Editing (RSE) mechanism. We apply RSE on top of the inference pipeline of a pre-trained InstructPix2Pix backbone~\cite{Instructpix2pix}. The relevance map computation [1] is adapted from prior work~\cite{watchyourstep}, while the anatomical pseudo masks and their integration with the relevance map to form the guidance map are proposed in this thesis. By restricting edits to instruction-relevant regions, RSE prevents unintended modifications and provides inherent interpretability via the visual guidance map.}
\label{fig:rse_mechanism}
\end{figure}

\label{Guidance Map}
region-specific editing prevents unintended modifications by precisely editing target regions. This method provides inherent interpretability by generating a guidance map, where the explanation directly reveals the decision process~\cite{attrinet}.  As shown in Figure~\ref{fig:rse_mechanism}, our approach combines model-derived relevance maps~\cite{watchyourstep} with dataset-derived pseudo masks to achieve precise and interpretable editing.


During inference, our region-specific editing process works as follows. First, given an input image \(I\) and an edit instruction \(T\), we encode \(I\) into the latent space and inject a fixed amount of Gaussian noise at a chosen diffusion timestep \(t_{rel}\). We then compute two noise predictions by feeding \((z_{t_{rel}}, I, T)\) and \((z_{t_{rel}}, I, T = \text{""})\) into \(\epsilon_\theta\). The relevance map (R) is obtained as the normalized absolute difference between these two predictions ($\epsilon_{I,T}(z_{t_{rel}})$,$\epsilon_{I}(z_{t_{rel}})$), which highlights regions that require modification according to the edit instruction~\cite{watchyourstep}:
\begin{equation}
\label{eq:relevance}
R_{x,I,T} = \text{normalize}\Bigl|\epsilon_{I,T}(z_{t_{rel}}) - \epsilon_{I}(z_{t_{rel}})\Bigr|.
\end{equation}

Next, to precisely localize anatomical regions associated with pathological findings, we incorporate expert-annotated bounding boxes from the MS-CXR dataset to create anatomically-aware pseudo mask. For each of the eight findings, we create individual pathology masks by taking the outer union of overlapping bounding box annotations, forming a pseudo set of eight masks. During inference, the final pseudo mask is generated by selecting each pathology mask from pseudo set that corresponds to the findings mentioned in the editing instruction, and then merging them into a single mask. For findings outside the eight annotated categories, we employ a $512 \times 512$ mask.

The final guidance map \(G\) is computed by multiplying \(R\) with the final pseudo mask \(M_{pseudo}\). \(\odot\) denotes element-wise multiplication:
\begin{equation}
\label{eq:guidance}
G = M_{pseudo} \odot R_{x,I,T},
\end{equation}
By this multiplication, the guidance map \(G\) effectively integrates (i) the model-identified regions to modify from the relevance map with (ii) the disease-related anatomical locations from the pseudo mask. The guidance map then represents pixel-wise information about precisely localized regions that align with editing instructions, serving as a visual explanation of the editing process for users.

We apply a threshold \(\tau\) on \(G\) to form a binary editing mask \(M_{x,I,T} = \mathbbm{1}(G \geq \tau)\). In each denoising iteration, we keep the unmasked regions identical to the input image by maintaining identical noise pixels in these areas throughout all steps. This approach prevents any modifications outside the region of interest to avoid unrelated spurious correlations. Furthermore, region-specific editing method supports user-defined mask specifications in place of \(M_{pseudo}\), enabling flexible control over editing process.

\section{Experiment}\label{c4}
\subsection{Implementation details}
We implement our framework based on the official codebase of InstructPix2Pix~\cite{Instructpix2pix}. Our model builds upon the pretrained InstructPix2Pix~\cite{Instructpix2pix}. The architecture employs a frozen CLIP text encoder and a U-Net backbone, processing triplet data ($I_{\text{past}}, I_{\text{cur}}, {T}$) from longitudinal chest X-ray pairs and editing instructions, as shown in the training pipeline of Figure~\ref{fig:training_framework}. Model trained on 8 A100 GPUs for 6,500 steps with learning rate $1 \times 10^{-4}$ and batch size 576. During inference, we set $\tau=0.1$ and fix $s_I=1.5$ and $s_{{T}}=7.5$.

\subsection{Baselines}\label{s4.2}
To rigorous evaluate the performance of InstructX2X, we compare it against four state-of-the-art baseline models encompassing both generative and editing approaches. Since some baselines were originally designed for \textit{de novo} generation rather than editing, we adapted their inference pipelines to ensure a fair comparison:

\begin{itemize}
    \item \textbf{RoentGen}~\cite{roentgen}: A latent diffusion-based model specialized for high-fidelity chest X-ray generation from radiology reports. As RoentGen is a text-to-image generation model rather than an image-to-image editor, we utilize the 'Impressions' section of the target radiology report as the input prompt to generate the counterfactual image from scratch, without conditioning on the source image.
    
    \item \textbf{LLM-CXR}~\cite{llm-cxr}: A comprehensive framework that utilizes a VQ-GAN and a text-only Large Language Model (LLM) for various chest X-ray vision tasks. Similar to RoentGen, we adapt LLM-CXR for our task by feeding the target findings description into its generation module to synthesize the corresponding chest X-ray.
    
    \item \textbf{BiomedJourney}~\cite{biomedjourney}: A counterfactual generation model that focuses on disease progression. It leverages GPT-4~\cite{achiam2023gpt} to synthesize a progression description based on the changes between prior and current reports. Following their official implementation, we use the GPT-4 generated progression summary as the instruction to transform the input image into the counterfactual target.
    
    \item \textbf{RadEdit}~\cite{radedit}: A unified framework for stress-testing biomedical vision models via image editing. RadEdit employs a multiple-mask strategy to maintain consistency during editing. For a fair comparison, we provide RadEdit with the same editing instructions and the same pathology-specific pseudo masks used in our Region-Specific Editing method.
\end{itemize}

\subsection{Evaluation Metrics}\label{s4.3}
We employ a comprehensive set of metrics to evaluate clinical accuracy, attribute preservation, and visual fidelity.

\subsubsection{CMIG Score}
To assess the trade-off between accurately modifying pathology and preserving patient identity, we adopt the Counterfactual Medical Image Generation (CMIG) score established by Cohen et al.~\cite{biomedjourney}. The CMIG score aggregates pathology accuracy ($a$) and feature retention ($f$) measurements using the geometric mean. This approach ensures robustness across varying scales and prevents the final score from being skewed by a single dominating factor. The CMIG score is defined as:
\begin{equation}
\text{CMIG Score} = \sqrt{\bar{a} \cdot \bar{f}} = \sqrt{\left(\prod_{i=1}^{n} a_i\right)^{1/n} \cdot \left(\prod_{j=1}^{m} f_j\right)^{1/m}}
\end{equation}
where $a_i$ represents accuracy metrics and $f_j$ represents feature retention metrics. We instantiate this score using three specific classifiers:

\begin{itemize}
    \item \textbf{Pathology Accuracy ($a_{path}$)}: We use the DenseNet-121 model from the XRV library~\cite{cohen2021xrv}, a state-of-the-art classifier for CheXpert findings. We compute the AUROC by comparing the predicted labels of the generated counterfactual images against the reference labels extracted from the target reports using the CheXpert labeler. We focus on the five most prevalent findings: Atelectasis, Cardiomegaly, Edema, Pleural Effusion, and Pneumothorax.
    
    \item \textbf{Race Preservation ($f_{race}$)}: To evaluate the preservation of demographic attributes, we employ a standard race classifier~\cite{gichoya2022}. We calculate the AUROC between the predicted race labels of the generated images and the self-reported race information from the MIMIC-CXR metadata.
    
    \item \textbf{Age Preservation ($f_{age}$)}: We use a deep neural network-based age regressor~\cite{ieki2022} to predict the patient's age from the generated image. We then compute the Pearson correlation coefficient between the predicted age and the patient's ground-truth anchor age recorded in the dataset.
\end{itemize}

\subsubsection{KL Divergence}
High classification scores alone can sometimes result from the model exploiting shortcuts or generating out-of-distribution features. To detect such anomalies, following the evaluation protocol of BiomedJourney~\cite{biomedjourney}, we measure the Kullback-Leibler (KL) divergence between the pathology probability distribution of real images ($P$) and synthesized images ($Q$). A lower KL divergence indicates that the generated images follow a clinically realistic distribution similar to real data:
\begin{equation}
D_{KL}(P || Q) = \sum_{i} P(i) \log \frac{P(i)}{Q(i)}
\end{equation}
where $P(i)$ and $Q(i)$ are the average predicted probabilities for the $i$-th pathology class across the test set for real and generated images, respectively.

\subsubsection{Fréchet Inception Distance (FID)}
To evaluate the perceptual quality and realism of the generated images, we employ the standard Fréchet Inception Distance (FID) metric. Following previous works in medical imaging, instead of the standard Inception-v3 network, we compute FID using using the feature embeddings from the pre-trained DenseNet-121~\cite{cohen2021xrv} to better capture domain-specific features of chest X-rays.

\begin{table}[t]
\centering
\caption{Quantitative comparison of InstructX2X with baselines and the ablation study. CMIG, KL divergence, and FID metrics are reported. \textbf{Bold} indicates the best performance among generated methods.}
\label{table 1}
\renewcommand{\arraystretch}{1.2}

\resizebox{\textwidth}{!}{%
    \begin{tabular}{l | c c c |c | c | c}
    \hline
    \multirow{2}{*}{\textbf{Model}}
    & \multicolumn{4}{c|}{\textbf{CMIG} (\(\uparrow\))}
    & \multirow{2}{*}{\textbf{KL} (\(\downarrow\))}
    & \multirow{2}{*}{\textbf{FID} (\(\downarrow\))} \\
    \cline{2-5}
    & \textbf{Patho (\(\uparrow\))} & \textbf{Race (\(\uparrow\))} & \textbf{Age (\(\uparrow\))} & \textbf{CMIG (\(\uparrow\))} &  & \\
    \hline
    Real Images (GT)
    & 84.66 & 99.56 & 88.83 & 90.99 & - & - \\
    \hline
    LLM-CXR
    & 55.82 & 48.62 & 0.10 & 52.06 & 60.03 & 51.42 \\

    RoentGen
    & 84.53 & 53.15 & 18.60 & 67.74 & 48.92 & 35.19 \\

    RadEdit
    & 82.80 & 80.33 & 48.24 & 80.14 & 38.85 & 26.31 \\

    BiomedJourney
    & 81.27 & 79.07 & 64.37 & 80.31 & 20.45 & 12.46 \\
    \hline
    InstructX2X (wo RSE)
    & 83.54 & 96.15 & 82.04 & 88.87 & 14.99 & \textbf{1.92} \\

    \textbf{InstructX2X}
    & 83.33 & 97.65 & 82.84 & \textbf{89.35} & \textbf{7.88} & 2.64 \\
    \hline
    \end{tabular}%
} 
\end{table}

\subsection{Results}\label{s4.4}
\subsubsection{Quantitative Analysis}
Table \ref{table 1} presents the quantitative comparison between InstructX2X and state-of-the-art baseline methods. Our proposed model, InstructX2X, achieves a comprehensive CMIG score of \textbf{89.35}, which is the highest among all generation methods and closely approaches the score of real ground-truth images (90.99). This indicates that our model successfully balances precise pathological editing with the preservation of patient identity. Specifically, InstructX2X demonstrates exceptional performance in preserving demographic attributes, maintaining Race preservation at \textbf{97.65} (vs. Real 99.56) and Age preservation at \textbf{82.84} (vs. Real 88.83), while simultaneously achieving a competitive Pathology modification score of 83.33.

In contrast, baseline methods exhibit significant performance imbalances. While RoentGen and RadEdit achieve relatively high pathology scores (84.53 and 82.80, respectively), they suffer from severe degradation in attribute preservation. For instance, RoentGen shows a catastrophic drop in Age preservation (18.60) and Race preservation (53.15), and RadEdit similarly struggles with Age preservation (48.24). This suggests that existing methods achieve disease modification at the cost of altering the patient's intrinsic identity, a critical failure for longitudinal medical analysis.

Furthermore, we analyze the reliability of the generated images using KL divergence. High pathology scores combined with high KL divergence, as seen in RoentGen (48.92) and LLM-CXR (60.03), imply that these models may be over-optimizing for specific features, resulting in a distribution that deviates significantly from real clinical data. InstructX2X achieves the lowest KL divergence of \textbf{7.88}, demonstrating that our generated images follow a distribution most similar to real data, ensuring that the high pathology performance is not a result of feature inflation but of realistic synthesis. Finally, in terms of visual quality, InstructX2X achieves an impressive FID score of 2.64, significantly outperforming baselines such as BiomedJourney (12.46) and RoentGen (35.19).

\subsubsection{Qualitative Analysis}

\begin{figure}[!ht]
    \centering
    \includegraphics[width=\textwidth]{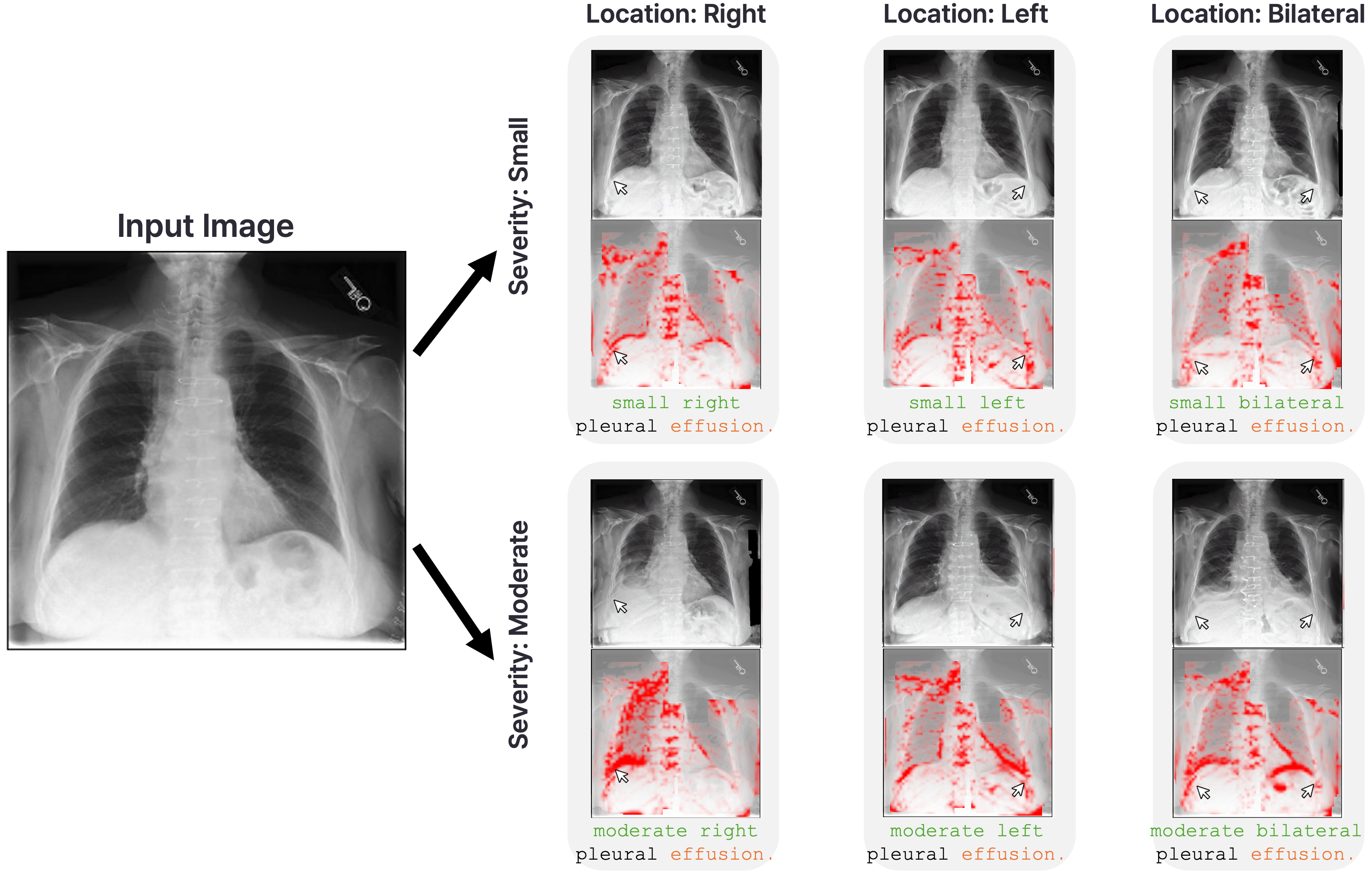}
    \caption{Demonstration of fine-grained controllable editing by InstructX2X. The model generates counterfactual images with varying severities (Small vs. Moderate) and precise anatomical locations (Right, Left, Bilateral) for pleural effusion. The guidance maps (red heatmaps) visualize the model's focus, confirming that edits are strictly confined to the instruction-specified regions and intensities.}
    \label{example}
\end{figure}

Building on the quantitative results, Figure~\ref{example} provides a qualitative assessment of InstructX2X's fine-grained controllability. Unlike existing methods that often generate generic disease patterns, our model demonstrates the ability to precisely manipulate both the \textbf{severity} and \textbf{anatomical location} of pathologies based on textual instructions.

As shown in Figure~\ref{example}, starting from a single input image, InstructX2X successfully generates six distinct variations of pleural effusion. Vertically, the model distinguishes between 'Small' and 'Moderate' severity levels; the generated opacities in the costophrenic angles are subtle in the 'Small' cases but become more pronounced and extensive in the 'Moderate' cases.

Horizontally, the model exhibits strict anatomical adherence. When instructed to edit the 'Right' or 'Left' side, the modifications—and their corresponding Guidance Maps—are exclusively localized to the specified lung field, leaving the contralateral side intact. In the 'Bilateral' case, the model correctly distributes the effect across both lungs. The Guidance Maps (red overlays) provide transparent verification of this process, confirming that the model's internal attention mechanism aligns perfectly with the spatial constraints of the medical instructions. This level of semantic and spatial control validates the effectiveness of our Region-Specific Editing approach in minimizing unwanted artifacts while maximizing clinical plausibility.

\subsection{Ablation Studies}\label{s4.5}
To validate the effectiveness of our proposed Region-Specific Editing (RSE) module, we conducted an ablation study comparing the base model (\textit{InstructX2X wo RSE}) with the full model (\textit{InstructX2X}). The bottom two rows of Table \ref{table 1} summarize the results.

The base model, which relies solely on global editing without region constraints, already achieves a respectable CMIG score of 88.87 and the best FID score of 1.92. This suggests that the foundational architecture and the quality of our instruction dataset are robust. However, the incorporation of the RSE module leads to critical improvements in model reliability and control.

By applying Region-Specific Editing, the KL divergence is drastically reduced from 14.99 to \textbf{7.88}. This reduction indicates that constraining the edit to relevant anatomical areas prevents the model from generating out-of-distribution artifacts, thereby aligning the generated pathology more closely with real-world distributions. Furthermore, RSE enhances the preservation of patient attributes, increasing the Race score from 96.15 to \textbf{97.65} and the Age score from 82.04 to \textbf{82.84}. Although there is a marginal increase in FID (from 1.92 to 2.64), this is a negligible trade-off considering the substantial gains in interpretability (via Guidance Maps) and the minimized risk of unintended background modifications. Consequently, the full InstructX2X model achieves the highest overall CMIG score of \textbf{89.35}, validating that region-specific control is essential for high-fidelity counterfactual medical image generation.

\section {Conclusion}
InstructX2X addresses two critical limitations of counterfactual medical image generation: unintended modification and insufficient interpretability. Our \textit{Region-Specific Editing} approach achieves precise feature  modification while preserving unrelated attributes, constraining the influence of spurious correlations during image generation. The \textit{Guidance Map} offers transparent visual explanations of the modification process, providing inherent interpretability rather than post-hoc explanations of uncertain reliability. By introducing the instruction-based \textit{MIMIC-EDIT-INSTRUCTION} dataset, we establish a more reliable foundation for future work. InstructX2X not only demonstrates state-of-the-art performance across multiple metrics, but it also confirm its clinical validity and explainability via radiologist assessments. These innovations collectively elevate counterfactual medical image generation for high-stakes clinical applications and AI model validation.

\appendix
\section{Example Appendix Section}
\label{app1}

Appendix text.


Example citation, See \cite{lamport94}.



\bibliographystyle{elsarticle-num}   
\bibliography{bib}                 
\end{document}